# End-to-End Learning vs. Modular Architectures: Comparative Insights into Autonomous Driving Systems

Kartik B. Kapse, kartik.b.kapse@gmail.com
(M.Eng.), AI Engineering of Autonomous Systems,
Dept. Electrical Engineering and IT, Technische Hochschule Ingolstadt
(Designed Framework: AD Architecture Selection Framework)

***Abstract-*** *Autonomous driving systems have become a central focus of intelligent transportation research, with End-to-End Learning and Modular Architectures offering two prominent design paradigms for their implementation. E2E Learning uses deep learning algorithms to map raw sensory inputs directly to driving actuators, providing a streamlined and adaptable solution. while Modular Architectures employ a pipeline-based approach, dividing the system into distinct subsystems for perception, cognition, planning, and control. This paper presents a comprehensive comparative analysis of these paradigms, focusing on their strengths, limitations, and trade-offs to provide insights into their suitability for various autonomous driving applications. The study evaluates key factors such as interpretability, scalability, robustness, and real-world applicability. While End-to-End Learning emphasizes simplicity and adaptability in dynamic environments, it lacks transparency and is highly dependent on large datasets. Conversely, Modular Architectures offer superior interpretability and task-specific optimization, but face challenges related to integration complexity and scalability. To address these limitations, hybrid approaches that combine the strengths of both paradigms have emerged, offering a promising direction for overcoming these challenges. Beyond this comparative synthesis, following work proposes a Four-Dimensional Architecture Selection Framework, comprising twelve binary criteria across safety, operating environment, data/computational resources, and deployment context, and validate it against ten published autonomous driving systems, correctly recommending 7/10 deployed architectures. This work synthesizes existing literature to highlight key trade-offs between the paradigms and identifies hybrid architectures as a promising direction for future research.*

## I. INTRODUCTION

Over the preceding decades, the domain of Autonomous Driving (AD) has garnered a substantial surge of interest from both academic institutions and industrial enterprises. The predominant catalyst underlying this heightened attention pertains to vehicular safety considerations. The National Highway Traffic Safety Administration (NHTSA) documented that approximately 94% of traffic accidents are attributable to human driver error [1]. An additional contributing factor relates to traffic flow optimization, wherein the substitution of human operators with AD systems yields enhanced traffic flow efficiency, conferring both economic and environmental advantages. The prospective benefits associated with fully realized AD systems appear considerable in magnitude, which substantiates the continued vitality of autonomous driving as an actively pursued research domain [2]. Autonomous Driving (AD) and Intelligent Vehicles (IVs) have garnered considerable interest from both the academic community and industrial sectors, owing to a broad spectrum of prospective advantages. Comprehensive surveys encompassing AD and IVs hold a pivotal role in consolidating research advancements, synthesizing the overarching trajectory of technological progression, and anticipating forthcoming developments [3].

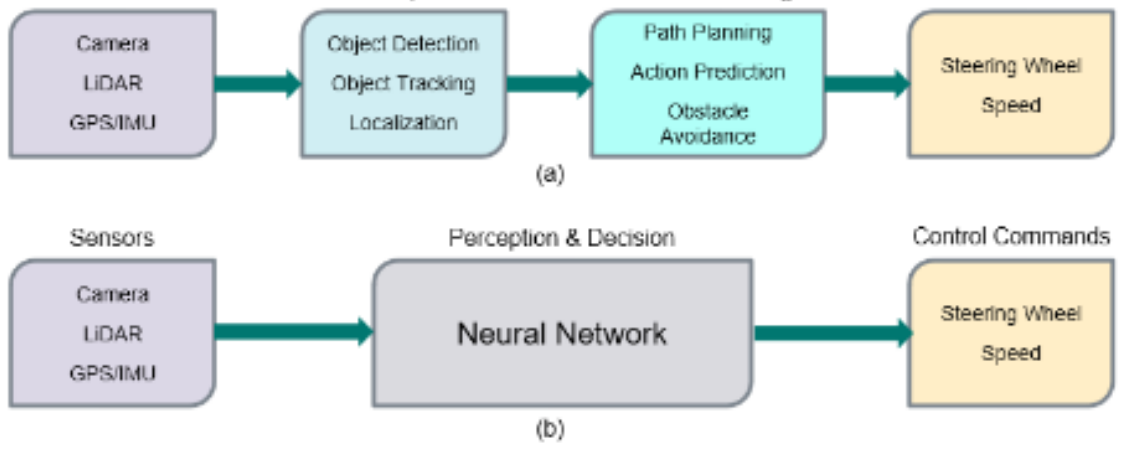

FIGURE 1.ARCHITECTURE OF: (A) A MODULAR APPROACH, AND (B) AN END-TO-END APPROACH. THE MODULAR ARCHITECTURE CONSISTS OF SEVERAL INTERCONNECTED MODULES, WHEREAS IN THE END-TO-END ARCHITECTURE, THOSE DIFFERENT MODULES ARE REPLACED BY A SINGLE, LEARNING-BASED MODULE

Autonomous Driving systems constitute inherently complex technological groups that merge a broad spectrum of constituent technologies, encompassing sensory apparatus, computational processing units, and software frameworks, among other components. Consequently, AD systems are necessitated to contend with an extensive array of operational challenges, including sensory inaccuracies, hardware reliability constraints, object detection limitations, and localization uncertainties. The conventional methodology employed to address this diversity of challenges involves the decomposition of the driving task into standardized functional modules such as object detection, localization, and path planning followed by the establishment of rule-based interfacing mechanisms to interconnect the discrete modules within a cohesive operational pipeline [4] [5] [6]. This approach is commonly called Modular Architecture (see Figure.1) and has attained widespread adoption across industrial applications. The interdependency and interconnectivity among discrete modules within a system represents a problem domain extensively investigated within the field of robotics. As a direct consequence of addressing such inter-module connectivity challenges, dedicated computational frameworks [2], have been developed, most notably the Robot Operating System (ROS) [7]

Modular architecture allows each module of a system to be developed independently, promoting seamless collaboration across the engineering team. By breaking down the autonomous driving system into specialized and well-researched components, such as localization, computer vision, and motion planning, it simplifies the overall task into manageable parts. One of the key advantages of this approach is its interpretability, since the system is divided into distinct modules identifying, and addressing the source of any malfunction becomes significantly easier. One significant drawback of modular systems is the challenge of developing and maintaining the interconnections between various modules. Different scenarios often demand unique module interactions, which can undermine the core principle of modularity. Moreover, these systems are susceptible to error propagation; for instance, a minor mistake, like misclassifying a traffic light, can cascade into severe consequences, such as incorrect path planning leading to potential collisions. Another limitation is their specialization, modules designed for specific tasks may struggle to adapt to unexpected situations or unusual conditions, limiting their ability to generalize effectively [2].

In recent years, end-to-end approaches have gained attention as an alternative solution for addressing the challenges of autonomous driving. These methods frame the task as a unified learning process, where the goal is to train a policy that directly translates sensor inputs into control commands. Compared to modular architectures, end-to-end systems are typically simpler, consisting of fewer components, and they better mimic the simultaneous perception and action characteristic of human driving. However, end-to-end systems face significant challenges, particularly their lack of interpretability [8]. It is often difficult to identify the source of errors or justify the decisions made by these models [9]. Historically, research in end-to-end approaches lagged behind modular systems, largely due to limited results and interest. But with the advancements in Artificial Intelligence, especially in Deep Reinforcement Learning (DRL) pioneered by organizations like DeepMind, end-to-end methods have started to demonstrate promising potential for autonomous driving applications [2].

In 1989 ALVINN (Autonomous Land Vehicle In a Neural Network) utilized an end-to-end learning approach for autonomous driving. This approach involved using a three-layer backpropagation neural network to map raw sensory inputs, such as images from a camera and a laser range finder, directly to steering commands. ALVINN demonstrated the ability to successfully navigate roads in simulated environments and under real-world conditions, highlighting the effectiveness of neural networks for road-following tasks in early autonomous systems. This method emphasized simplicity by bypassing traditional modular pipelines, which separate perception, planning, and control, and instead learned driving behavior as a unified task. However, the approach also highlighted challenges such as the dependency on diverse and representative training data for robust real-world performance [10].

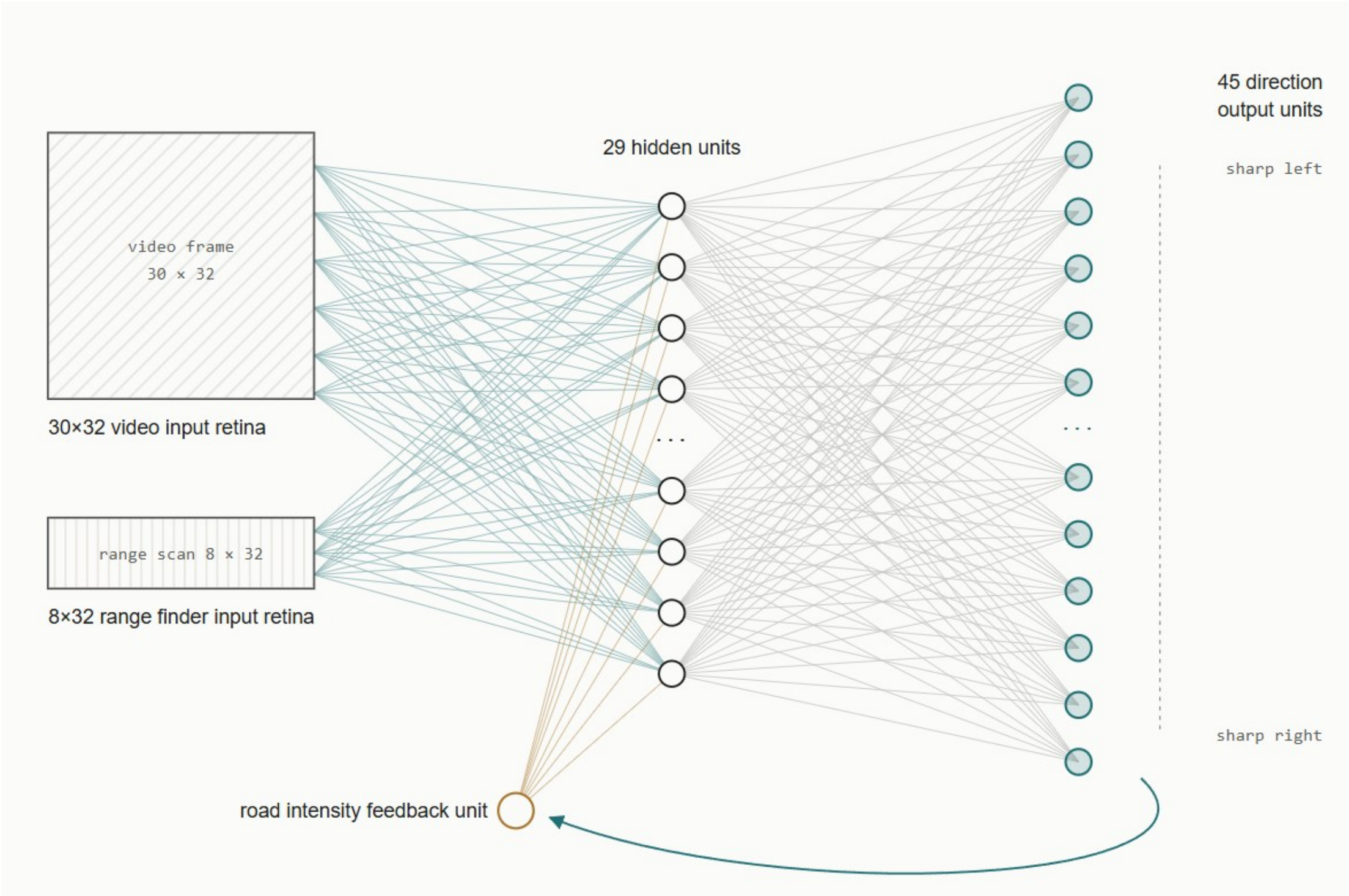

**FIGURE 2. ALVINN ARCHITECTURE. ADAPTED FROM [10]**

The ALVINN (Autonomous Land Vehicle In a Neural Network) architecture (Figure 2) processes sensory inputs from a vehicle's camera and laser range finder. Its input layer includes two "retinas": a video retina with 30x32 units capturing road images using the blue color band for high contrast and a range retina with 8x32 units mapping proximity data from the laser. An additional feedback unit tracks road intensity changes. These inputs connect to a hidden layer of twenty-nine units and an output layer of forty-six units. The output predicts turn curvature, guiding the vehicle towards the road center, with outputs ranging from sharp left to sharp right turns [10].

End-to-end approaches employ two primary learning methodologies: Imitation Learning (IL) and Reinforcement Learning (RL). Imitation Learning involves training a model to replicate human behavior by observing human actions, functioning as a supervised learning method [11]. NVIDIA demonstrated the potential of this approach by using a convolutional neural network (CNN) to predict vehicle steering angles with remarkable accuracy [12]. One significant advantage of IL is its reliance solely on Deep Learning (DL), where the model's parameters are fine-tuned to minimize the difference between its behavior and that of a human driver. However, scaling autonomous driving systems using IL presents challenges, as it is nearly impossible to encompass every possible driving scenario during the training phase [2].

Reinforcement Learning (RL) focuses on training a policy that maximizes cumulative rewards by allowing an autonomous system to interact with its environment. A specialized form, Deep Reinforcement Learning (Deep RL), combines Deep Learning (DL) with RL techniques. While there are technical distinctions between RL and Deep RL, this discussion will use the terms interchangeably to highlight the differences between RL and Imitation Learning (IL). Unlike IL, RL does not require human driving data. Instead, the system learns optimal behaviors by interacting with the environment and adjusting to maximize rewards. This online training capability allows RL models to explore and train simultaneously, offering a significant advantage over IL methods. However, RL is less data-efficient during training, making it computationally intensive. To address these limitations, researchers such as Liang et al. have proposed hybrid methods that combine the strengths of IL and RL. For instance, an initial IL model can be trained using labelled data, followed by online RL-based policy refinement to improve performance. A critical component of RL algorithms is the reward function, as it directly shapes the agent's learning and behavior. Common examples include rewarding progress toward a goal or penalizing collisions, ensuring the agent develops safe and efficient navigation strategies [2].

Designing systems for autonomous driving has various challenges which requires its architectural designs to be examined carefully. End to end learning maps directly from sensory input to driving outputs using deep neural networks. While end-to-end learning offers simplicity, it does not allow the user to trace back the cause of a bad decision since it acts as a black box. Also, it requires large and varied datasets to work effectively, but it does not work on rare or wild situations. Modular architectures break down the driving process into subsystems like perception and planning which enables task-specific robustness and interpretability. But they have their own difficulties like integration issues and error propagation, which means errors occurring in one module propagate through other modules and make the whole system fail. The trade-offs between adaptability, transparency, and reliability show the importance of seeking alternatives. A combination of the advantages of both paradigms or the application of Explainable AI (XAI) techniques can lead to scalable, robust, and trustworthy autonomous systems capable of a dynamic real-world environment. [13]

The field of autonomous driving has witnessed extensive research on both end-to-end learning and modular architectures, with each paradigm offering unique advantages and limitations. However, there is a clear research gap in comprehensive studies that directly compare these approaches across critical dimensions such as scalability, robustness, and interpretability [14]. This lack of comparative analysis creates challenges in identifying the most suitable framework for specific real-world applications. Moreover, hybrid models, which blend the strengths of both paradigms, have emerged as a promising alternative, yet their potential remains underexplored. These hybrid approaches could provide a balance between the interpretability and modularity of traditional systems and the adaptability and efficiency of end-to-end learning models. Despite this promise, significant challenges remain, including error propagation in modular systems, data dependency in end-to-end models, and the absence of standardized benchmarks for fair evaluation. Addressing these gaps is crucial to advancing the field and enabling more reliable, scalable, and effective autonomous driving systems. [15] [16] [14]

To help bridge this gap, this paper introduces a Four-Dimensional Architecture Selection Framework for Autonomous Driving System Design, a practical decision-support tool intended to guide practitioners toward the most appropriate paradigm Modular, End-to-End Imitation Learning, End-to-End Reinforcement Learning, or Hybrid for a given deployment context. The framework organizes the architecture selection problem into four sequential dimensions: safety and regulatory compliance, operating environment, data and computational resources, and deployment and operational context, operationalized through twelve binary criteria (C1-C12) that map directly onto paradigm recommendations. To assess its practical validity, the framework is applied retrospectively to ten representative autonomous driving systems spanning three decades of development, from ALVINN (1989) to CaRINA Agent (2023), comparing its predicted architecture against each system's actual deployed design.

This paper aims to provide a comparative analysis of End-to-End Learning and Modular Architectures, focusing on key evaluation factors such as interpretability, scalability, robustness, and real-world applicability. The remainder of this paper is organized as follows: Section II delves into the foundational paradigms of autonomous driving, discussing End-to-End Learning, Modular Architectures, and Hybrid Approaches, along with their methodologies and real-world examples. Section III introduces the proposed Four-Dimensional Architecture Selection Framework and presents a comprehensive comparative analysis exploring key aspects such as technical architecture, performance metrics, scalability, and practical applicability, before expanding on the discussion of trade-offs, highlighting potential applications, and examining emerging trends, with a focus on hybrid solutions as a bridge between the two paradigms. Finally, Section IV concludes the study by summarizing the findings, emphasizing critical insights, and offering recommendations for future research and practical implementation.

The key contributions of this paper are as follows:

- A comprehensive comparative synthesis of End-to-End Learning, Modular, and Hybrid architectures for autonomous driving, evaluated across interpretability, scalability, robustness, and real-world applicability.

- A Four-Dimensional Architecture Selection Framework, comprising twelve binary criteria (C1–C12) organized across safety and regulatory compliance, operating environment, data and computational resources, and deployment and operational context, that maps design requirements onto Modular, End-to-End Imitation Learning, End-to-End Reinforcement Learning, or Hybrid recommendations.
- An empirical validation of the proposed framework against ten representative real-world autonomous driving systems (1989–2023), demonstrating correct or partial architecture recommendations for seven of ten cases and identifying the framework's scope boundaries.
- A synthesis of emerging hybrid design trends and an outline of open research directions, including standardized cross-paradigm benchmarking, formally verifiable hybrid safety properties, and sim-to-real transfer for end-to-end policies.

This review draws on literature identified through IEEE Xplore, arXiv, Google Scholar, and MDPI, using search terms including "end-to-end autonomous driving," "modular autonomous driving architecture," "imitation learning autonomous vehicles," "reinforcement learning autonomous driving," and "hybrid autonomous driving architecture," combined with terms for specific subsystems (perception, planning, motion prediction, sensor fusion). The search covered work published between 1989 (marking the introduction of ALVINN) and 2025. Sources were prioritized by citation count and venue, favouring peer-reviewed journal and conference publications (e.g., IEEE Access, IEEE Transactions on Intelligent Vehicles, ICRA, ECCV) alongside high-citation arXiv preprints where peer-reviewed versions were unavailable or the preprint represented the most current version of the work. This paper does not follow a formal systematic review protocol (e.g., PRISMA); rather, it adopts a narrative synthesis approach appropriate for comparing architectural paradigms across a heterogeneous body of work spanning three decades, sensor modalities, and evaluation benchmarks

## II. Autonomous Driving Paradigms

### *A. Autonomous Driving Paradigms Overview:*

Autonomous driving systems are generally designed using one of two key paradigms: end-to-end learning or modular architectures, each offering distinct methodologies and advantages. End-to-end learning aims to directly map raw sensory inputs, such as images from cameras or data from LiDAR sensors, into vehicle control commands like steering, throttle, and braking. This approach uses deep learning to integrate perception and action into a single streamlined framework, mimicking human driving behavior. The simplicity and adaptability of this paradigm make it suitable for complex and dynamic environments. In contrast, modular architectures decompose the autonomous driving task into a sequence of well-defined modules, enabling a step-by-step approach to processing and decision-making. These modules typically include:

1. Sensing: Capturing raw environmental data through sensors like cameras and LiDAR.
2. Perception: Interpreting sensor data to identify objects, lanes, and other relevant features.
3. Prediction/Decision Making: Anticipating the behavior of other entities on the road.
4. Planning: Determining an optimal and safe trajectory for the vehicle.
5. Control: Executing the planned trajectory through vehicle actuators.

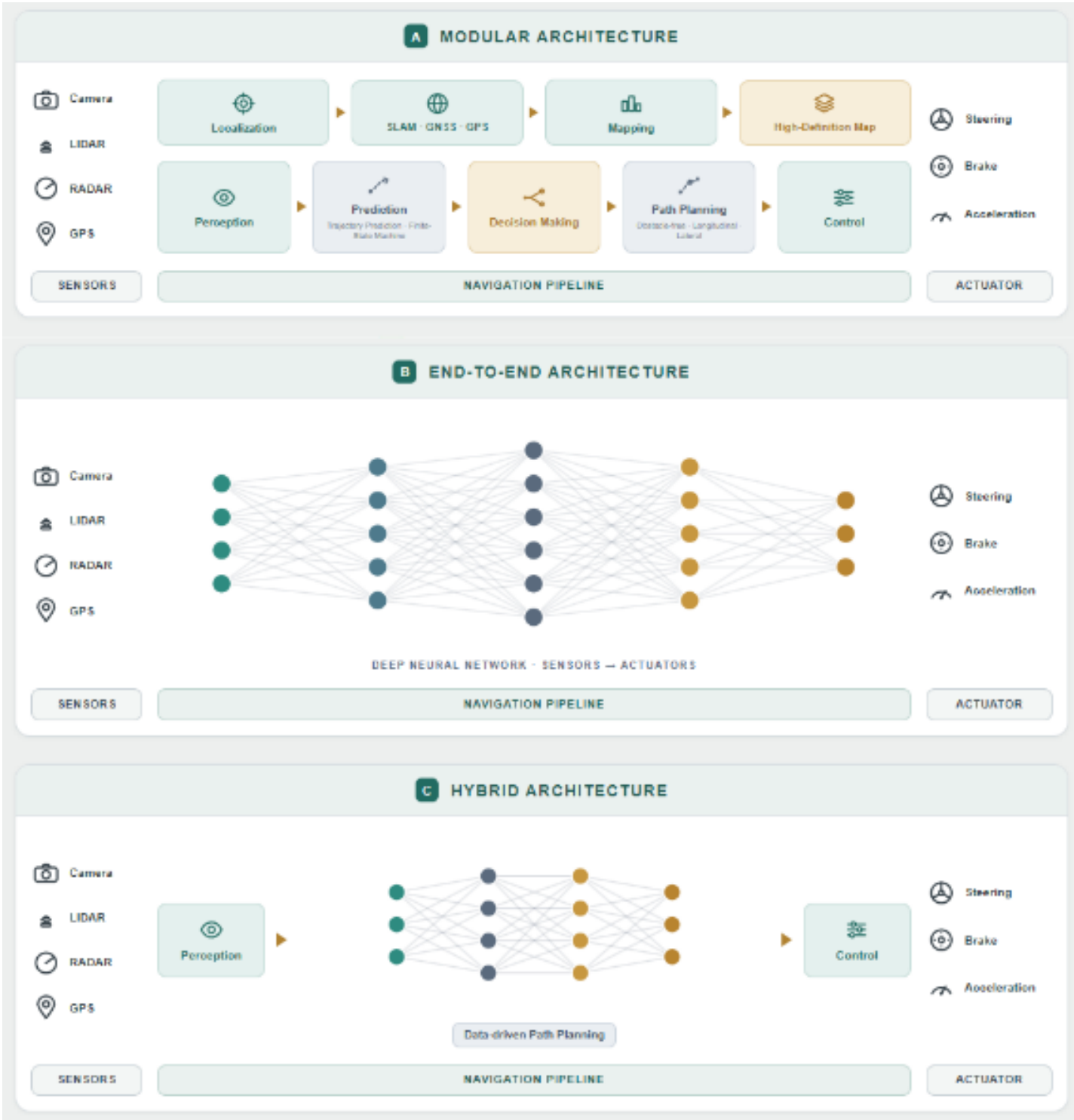

**FIGURE 3. ARCHITECTURE ILLUSTRATION, A) MODULAR, B) END-TO-END, AND C) HYBRID. REPRODUCED FROM [15] UNDER CC BY 4.0**

This modular breakdown facilitates specialized optimization for each stage, improving system interpretability and maintainability. The hybrid model (Figure 3) seeks to overcome the limitations of both modular and end-to-end approaches by combining their strengths. By merging the interpretability and deliberate decision-making of modular systems, especially effective when working with map-based data, with the adaptive learning capabilities of end-to-end models, which excel at processing complex sensory inputs, hybrid models offer a more balanced solution. This integration enhances system robustness by uniting reactive and deliberative behaviors, improving flexibility, adaptability, and overall performance in diverse driving scenarios [15].

In this section, we first discuss the modular paradigm, exploring its core components and methodologies. Subsequently, we provide an overview of end-to-end learning, highlighting its innovative approach to integrating sensory data and control signals. These insights set the stage for understanding how the strengths of both paradigms can be combined to address current limitations and drive advancements in autonomous driving technology. [16]

*B. Modular Architecture*

Modular systems simplify the complex task of autonomous driving by breaking it down into smaller, manageable modules, each addressing a specific sub-task [17]. These modules are often further subdivided into components, with well-defined intermediate representations and interdependencies. This modular breakdown improves interpretability, making it easier to trace errors to specific modules in the event of unexpected behavior [8]. Additionally, modularization facilitates leveraging existing knowledge and methods from fields like computer vision and robotics [18]. Another significant benefit of modular systems is their flexibility: algorithms within modules can be updated or replaced without requiring extensive changes in other parts of the system [16].

However, modular systems face notable challenges. They often struggle with generalization to diverse scenarios, which impacts overall robustness [8] [18]. Since modules rely on human-engineered input-output representations and heuristics, they may overlook contextual information critical for downstream tasks. For example, object detection in perception modules typically outputs only a bounding box and class label, omitting scene context that could enhance decision-making. Modular systems are also prone to complexity, with multiple models and intermediate representations adding to computational overhead. Many models, especially those used in perception, depend heavily on supervised data for training, which creates a bottleneck defined by human-provided information [17] [16]. Furthermore, errors in one module can propagate through the system, leading to compounded issues downstream.

The following discussion will explore the five primary modules of modular systems, the techniques they employ, and the challenges associated with their implementation.

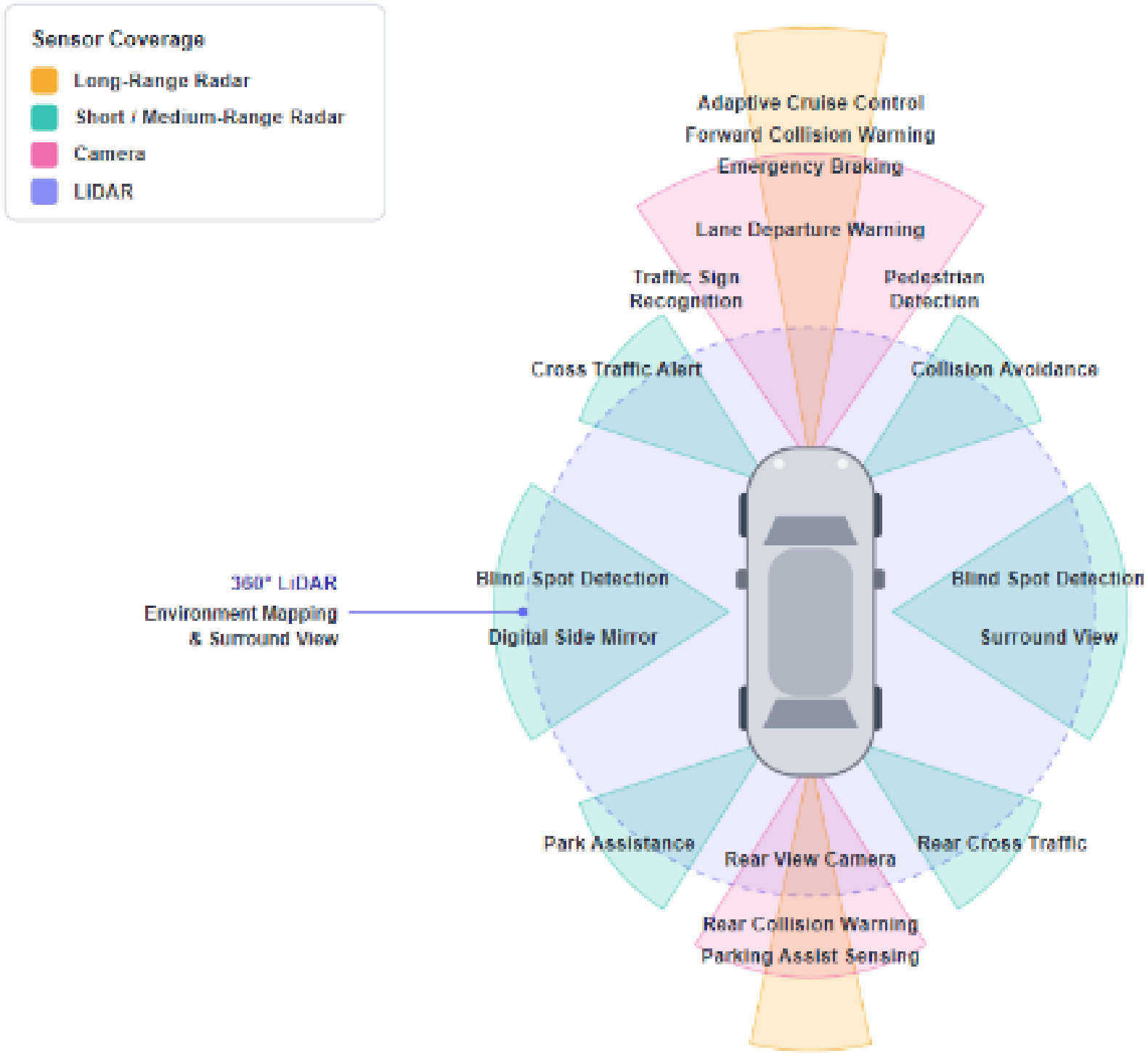

**FIGURE 4. EXAMPLE SENSOR CONFIGURATION IN AN AUTONOMOUS VEHICLE [19] [16]**

*a) Sensing*

The sensing module of an Autonomous Driving System (ADS) employs a combination of advanced sensor technologies to provide high-resolution environmental data essential for vehicle operation. Cameras capture detailed color and texture information, while LiDARs[1] create accurate 3D maps and measure object depth. Radars[2] reliably detect object velocity and range, even in adverse weather conditions. GPS delivers precise positioning data, and IMUs[3] track motion by measuring acceleration and angular velocity. Each sensor has unique strengths and limitations, so ADS typically integrates data from multiple sensors using sensor fusion techniques. This fusion enhances reliability by compensating for individual sensor weaknesses, creating a comprehensive view of the driving environment. The resulting data enables the ADS to perceive surroundings accurately, plan safe paths, and make critical driving decisions in real-time. [19]

*b) Perception*

The perception module is a critical component of Autonomous Driving Systems (ADS), tasked with processing data from various sensors to construct an accurate and meaningful representation of the driving environment. This module typically employs advanced computer vision techniques, which have seen rapid advancements due to the integration of deep learning. Core perception tasks include object detection, object tracking, and lane detection, each addressing specific challenges to ensure safe and efficient vehicle operation [20] [18] [17].

[1] LiDAR is a remote sensing technology that is used for measuring distances

[2] Radar is a technology that uses radio waves to measure the distance, angle, and velocity of objects. It works on the principle of electromagnetic radiation that can be used in multiple frequency bands. (e.g. 24 GHz, 77 GHz, 79 GHz)

[3] An Inertial Measurement Unit (IMU) is an electronic device that can measure a body's force, angular rate, and magnetic field. IMU devices usually consist of three accelerometers, gyroscopes, and magnetometers, one for each orthogonal X, Y, and Z axes.

1. Object detection

Object detection involves identifying and localizing objects of interest within a single camera image. These objects may be static (e.g., traffic lights, road signs) or dynamic (e.g., other vehicles, pedestrians). It is one of the most critical components in ADS due to its direct impact on safety. For instance, a failure in detecting a vehicle can propagate errors to downstream modules like planning and control, potentially causing collisions [17].

- State-of-the-Art Techniques:
    - Two-stage detectors (e.g., Faster R-CNN, Mask R-CNN) utilize a Region Proposal Network (RPN) to generate regions of interest, followed by classification and bounding-box regression [21] [22].
    - One-stage detectors (e.g., YOLO, SSD) bypass the RPN and directly perform regression on densely sampled locations, prioritizing speed over accuracy [23] [24] [25] [26].
        - A common challenge in object detection is balancing the accuracy-latency trade-off, where more complex models tend to deliver higher accuracy but with increased computational cost and latency.

2. Object Tracking

Object tracking involves maintaining consistent identifiers for objects across video frames, enabling the generation of object trajectories. This is particularly useful for trajectory prediction in ADS.

- Approaches:
    - Traditional methods, like SORT, employ techniques such as the Kalman Filter and the Hungarian algorithm to maintain bounding boxes and IDs [27].
    - Deep learning-based trackers deliver state-of-the-art performance but often at the expense of higher latency [16].
- Despite their lower accuracy, lightweight trackers like SORT remain relevant for real-time applications due to their minimal computational overhead [18].

3. Lane detection

Lane detection identifies lanes and lane segments in camera images, providing crucial information for understanding road semantics and vehicle navigation. This task is challenging due to variations in road topology, marking styles, and lane semantics [17].

- Techniques:
    - Traditional methods rely on color, texture, and gradient-based geometric modeling or energy minimization algorithms [28].
    - Deep learning approaches, such as segmentation models and CNN-RNN hybrids, offer improved robustness but come with computational demands [29] [30].

4. Performance Considerations and Challenges

The accuracy-latency trade-off is a significant concern in perception tasks. Complex deep learning models, while highly accurate, introduce latency that can be detrimental to real-time decision-making. Conversely, traditional methods, although less accurate, excel in delivering rapid results. This trade-off must be carefully managed to meet the strict runtime requirements of ADS. For example, in urban scenarios, where vehicles may travel at 40 km/h, the control signal must be ready in less than 90 milliseconds to ensure safety [31].

Errors in the perception module propagate through the autonomous driving pipeline, leading to suboptimal planning and control decisions, potentially resulting in safety risks. Additionally, achieving the balance between detection speed and accuracy is critical for ensuring that the ADS can operate reliably in dynamic and time-sensitive environments.

In summary, perception forms the backbone of autonomous driving, with advancements in computer vision and deep learning continuing to refine its capabilities. However, the challenges of sensor integration, trade-offs between accuracy and latency, and real-time processing constraints highlight the need for ongoing innovation in this domain.

*c) Prediction*

The prediction module leverages environmental data furnished by the perception system to evaluate the behavioral dynamics of proximate vehicles and pedestrians, thereby facilitating the assessment of potential risk factors inherent within the prevailing driving scenario [17]. It does so by forecasting the future positions of these agents, a process known as motion trajectory prediction. Object tracking is essential for this task, as it monitors each agent's past states over a specific time period. This contextual information facilitates the estimation of vehicular velocity and lane occupancy positions. Beyond historical tracking data, the system additionally assimilates scene-level contextual details encompassing 3D point cloud representations derived from LiDAR sensors and bird's eye view perspectives of the surrounding environment to augment predictive accuracy [32]. A linear regression model serves as the baseline predictive framework, operating under the assumption that each agent maintains a constant velocity, without accounting for surrounding scene context or driver behavioral tendencies. Analogous to the SORT tracker, linear regression exhibits comparatively diminished predictive performance relative to deep learning models, yet offers substantially superior computational efficiency. For a comprehensive examination of deep learning-based prediction methodologies, readers are directed to [32].

Prediction constitutes an inherently challenging yet critically indispensable component of autonomous driving systems, providing essential informational inputs that enable the vehicle to execute proactive behavioral responses, such as lane transitions or intersection navigation. The computational complexity of this task escalates considerably upon the involvement of multiple dynamic agents within the operational scene, as the behavioral trajectories of all constituent agents necessitate simultaneous monitoring and predictive inference.

*d) Planning*

The planning module bears responsibility for generating a structured sequence of temporal waypoints that collectively define the intended trajectory for the ego-vehicle. Operational inputs to this module comprise the predicted behavioral trajectories of surrounding agents, supplied by the prediction module, in conjunction with localization and cartographic data originating from the constituent components of the perception module. This computational process is conventionally organized into three hierarchical stages, namely path planning, behavior planning, and motion planning.

Path planning: Path planning in autonomous vehicles denotes the computational methodology through which self-navigating systems determine the optimal and safest trajectory from a given origin to a specified target destination. This process encompasses obstacle avoidance, compliance with established traffic regulations, and accounting for the dynamic characteristics of the surrounding operational environment. Leveraging the vehicle's instantaneous positional data, its intended destination, and information acquired from cartographic resources and sensory apparatus, the path planning algorithm synthesizes a preliminary trajectory connecting the vehicle to its objective while circumventing potential impediments. The underlying objective function incorporates parameters such as traversal distance, passenger comfort, operational safety, energy consumption, and probabilistic collision risk assessment, with the overarching aim of identifying a minimum-cost trajectory. Within Modular architectures, path planning predominantly relies upon rule-based design paradigms, though such methodologies frequently demonstrate inadequate adaptability when confronted with the extensive variability of real-world driving scenarios [33] [16].

Behavior planning occupies the second position within the sequential processing hierarchy. Subsequent to the generation of the route plan, the Behavior planner determines the appropriate high-level action directives to be executed by the ego-vehicle. Such actions conventionally encompass lane maintenance, preparatory lane-change maneuvers, or adherence to designated waypoint sequences. Owing to the inherent stochasticity and uncertainty characterizing the operational driving environment, behavior planning frequently employs probabilistic modeling frameworks like Markov Decision Processes (MDP) and Partially-Observable MDPs (POMDPs) to facilitate informed decision-making across a spectrum of plausible environmental outcomes and situational scenarios [34] [16].

Finally, motion planning alternatively designated as trajectory planning employs the selected high-level behavioral directive to synthesize a safe, comfortable, and dynamically feasible trajectory toward the intended goal destination. Prevalent computational techniques within this domain encompass graph search methods [16] and incremental search methods [35]. Graph search methods construct a graphical representation of the vehicle's configuration space and systematically seek the minimum-cost traversal path [16]. Incremental search methods, conversely, progressively construct a tree structure of reachable states emanating from the vehicle's initial positional configuration, subsequently identifying the optimal branch within the generated tree. A detailed overview of motion planning methods is provided in [6].

*e) Control*

Control pertains to the ensemble of systems and computational algorithms responsible for translating high-level decisional outputs and pre-established trajectories into precise operational commands executed by the vehicle's actuation mechanisms. These regulatory frameworks ensure adherence to the designated route, facilitate dynamic speed modulation, and enable adaptation to diverse driving scenarios while preserving operational safety, vehicular stability, and passenger comfort. Fundamentally, control systems serve as the intermediary layer bridging the cognitive decision-making pipeline and the physical locomotive behavior of the vehicle. The constituent components of such systems encompass actuation mechanisms, vehicular dynamic modeling, longitudinal speed regulation, maneuverability characteristics, stability maintenance, seamless transitional responses, and the optimization of human ergonomic comfort [33].

*C. End-to-End Learning*

Attention is now directed toward End-to-End methodologies within autonomous driving control systems. In contrast to decomposing the driving task into discrete functional modules, End-to-End approaches establish a direct mapping between raw sensory inputs and executable control commands. Such systems are characteristically less architecturally complex than Modular architectures, encompassing fewer constituent components and comparatively simplified model structures. [8]. Nevertheless, this architectural simplicity is accompanied by a trade-off in interpretability, as elucidating the rationale underlying the model's behavioral outputs becomes increasingly challenging, particularly upon the manifestation of operational anomalies. This stands in notable contrast to modular systems, wherein the localization and attribution of failure to a specific constituent module is considerably more tractable.

End-to-End approaches can be broadly classified into two principal categories: imitation learning and reinforcement learning. Each of these methodological categories will be examined in greater depth in the subsequent sections.

*a) Imitation Learning*

In imitation learning an autonomous agent acquires task-execution capabilities through the replication of behavioral patterns demonstrated by a human expert or a pre-established reference model. Rather than formulating an explicit reward function or deriving an optimal policy through conventional means, the agent observes and systematically reproduces the action sequences exhibited by the expert demonstrator. This paradigm offers several notable advantages, including the leveraging of domain-specific human expertise to address inherently complex tasks, while frequently yielding safer and more interpretable behavioral outputs in contrast to reinforcement learning frameworks, wherein agents depend upon iterative trial-and-error exploration. Nevertheless, imitation learning is accompanied by certain limitations, notably its susceptibility to propagation of inaccuracies present in the expert's demonstrations and its inherent constraint in identifying strategies that transcend the boundaries of the demonstrated behavioral repertoire. Throughout the training phase, the agent's objective is to establish a functional mapping between input observational data and the corresponding actions of the expert demonstrator. This is conventionally accomplished via supervised learning mechanisms, wherein the model is optimized to minimize the discrepancy between its predicted action outputs and those documented within the expert's demonstration dataset. Analogously, during the evaluation phase, the agent's generalization capability is assessed on previously unseen data to quantify the efficacy with which it reproduces the expert's behavioral policy. This evaluative process may entail a comparative analysis between the agent's generated actions and those of the expert, or alternatively, an assessment of task completion performance under varied operational conditions.
[33].

Imitation learning can be further categorized into two main types: behavior cloning and inverse reinforcement learning. However, before delving into these categories, it is important to understand the problem formulation.

*Problem Formulations:* In imitation learning, a dataset $\mathbb{D}$ consists of multiple trajectories, where each trajectory is a sequence of state-action pairs (S, A). The action *A* represents the decision made by the expert at a particular state *S*, guided by the expert's policy $\pi_E$ [36] [37]. The goal of imitation learning is to optimize and learn a policy $\pi^*$: $S \rightarrow A$ that closely mimics the expert's policy $\pi_E$, ensuring the following condition is satisfied:

$$\pi^* = argmin_{\pi} \mathbb{D}(\pi_E, \pi) \qquad (1)$$

Where $\mathbb{D}$ represents the similarity between current and expert policies.

*Behavior Cloning:* One of the earliest studies on behavior cloning was conducted with the ALVINN system [10], which used a fully connected neural network to perform lane-following tasks. While modern approaches now use deep neural networks instead of simple fully connected layers, the underlying principle of behavior cloning remains unchanged [12] [38]. In behavior cloning, the problem is framed as a supervised learning task, aiming to align the learned policy $\pi_\theta$ with the expert policy $\pi_E$. This alignment is achieved by minimizing the L2 loss:

$$min_{\theta} \mathbb{E} \|\pi_\theta - \pi_E\|_2 \qquad (2)$$

As the name implies, behavior cloning methods are limited to imitating the actions seen in the training dataset. They cannot generalize effectively to new, unseen scenarios. This limitation often results in a covariate shift; a specific type of dataset shifts where the distribution of input data changes between the training and live environments [39]. Covariate shift leads to compounding errors when the system is exposed to new states. Subsequent research has addressed this issue by using data augmentation and regularization techniques. These methods introduce noise into the actions and learning process to enhance robustness and improve generalization [40].

*Inverse Reinforcement Learning (IRL):* Inverse Reinforcement Learning (IRL) enables an autonomous agent to recover the latent reward function governing an environment through systematic observation of expert demonstrations or behavioral trajectories. In contrast to conventional reinforcement learning, which is oriented toward the maximization of a predefined reward function, IRL endeavours to reconstruct the underlying reward structure that most plausibly accounts for the expert's sequential decision-making. This paradigm proves particularly advantageous in scenarios where the explicit manual specification of a reward function is inherently intractable yet can be inferred from observed expert behavioral patterns.

$$max_{\theta} \mathbb{E}_{\pi E}[G_t | r_\theta] - \mathbb{E}[G_t | r_\theta] \qquad (3)$$

Inverse Reinforcement Learning (IRL) operates upon a dataset comprising demonstrative trajectories sequential state-action pairs that encapsulate the navigational behavior of an expert agent within a given environment. Throughout this process, the agent performs systematic feature extraction from the state space to discern salient characteristics of the operational environment, thereby facilitating generalization across heterogeneous state configurations. The fundamental objective of IRL is to recover a latent reward function that attributes elevated values to expert-consistent behavioral patterns while assigning diminished values to actions deviating from expert-level performance. This problem is conventionally formulated as a constrained optimization task. Upon successful recovery of the reward function, the agent is equipped to employ established reinforcement learning methodologies such as Q-learning or policy gradient-based approaches to derive an optimal behavioral policy that maximizes the reconstructed reward signal.

IRL faces three primary challenges in autonomous driving. First, expert human driving data is inherently noisy and suboptimal, drivers make minor errors and stylistic choices that do not reflect globally optimal behavior, leading to ambiguous reward inference. Second, in the high-dimensional state space of AD (LiDAR point clouds, camera images), reward function recovery requires enormous data volumes, making IRL computationally prohibitive compared to behavior cloning. Third, rewards learned in one environment (e.g., highway driving) generalize poorly to structurally different scenarios (e.g., urban intersections), limiting cross-domain applicability.

IRL is a powerful tool but requires careful consideration of these challenges to ensure effective and reliable learning outcomes.

b) Reinforcement Learning

Reinforcement Learning (RL) enables an autonomous agent to acquire decision-making capabilities and develop behavioral strategies within a given environment through the maximization of a cumulative expected reward. Within the context of autonomous systems, RL facilitates task acquisition and sequential decision-making via iterative experiential exploration, circumventing the necessity for explicit procedural programming.

Within the RL framework, the agent constitutes the learning entity that engages in continuous interaction with an external environment to accomplish predefined objectives. The environment occupies a distinct state at each discrete time step, representative of its instantaneous configuration. The agent perceives this state and subsequently selects an appropriate action from its available action space. The executed action exerts influence upon the environment, inducing a transition to a subsequent state while concurrently generating a reward signal a numerical scalar quantifying the immediate efficacy or inadequacy of the undertaken action. The agent's policy represents a strategic mapping from observed environmental states to corresponding actions. The overarching objective of the agent is to converge upon an optimal policy that maximizes the accumulated reward over an extended temporal horizon. The value function constitutes another fundamental component of the framework, providing an estimation of the total expected cumulative reward attainable from a given state, thereby endowing the agent with the capacity for informed decision-making. Through iterative engagement with the environment, the agent employs RL algorithms to progressively refine its policy or value function approximation. This optimization process inherently balances the trade-off between exploration the systematic investigation of novel actions to uncover superior behavioral strategies and exploitation the utilization of previously acquired knowledge to maximize immediate reward accumulation

Managing this trade-off is one of RL's central challenges. Strategies like $\epsilon$-greedy and SoftMax help tune exploration and exploitation. Agents generally explore more during the early stages of training when little is known about the environment.

For sequential decision-making problems, RL often uses Markov Decision Processes (MDPs). An MDP consists of:

- $S$: A set of states
- $A$: A set of actions
- $T$: A transition function that defines the likelihood of moving from one state to another based on an action
- $R$: A reward function providing feedback for actions

Reinforcement learning approaches can be categorized into value-based, policy-based, and actor-critic methods. Despite its advantages, RL also faces challenges like efficient exploration, handling long-term rewards, and ensuring safe learning in real-world applications.

*1. Value-Based Reinforcement Learning (RL):* Value-based RL focuses on estimating the desirability of states or state-action pairs to guide decision-making. This desirability, or *value*, represents the expected cumulative reward for the agent when in a specific state or taking a particular action.

- *State-Value Function (V(s)):* Represents the expected reward the agent will receive from a state '*s*' while following a specific policy.
- *Action-Value Function (Q(s, a)):* Represents the expected reward from taking action '*a*' in state '*s*' and continuing to follow the policy.

The Bellman equation is central to value-based RL, helping update value estimates based on future rewards. The goal is to approximate the optimal value function, which provides the best possible outcomes under the ideal policy.

A popular algorithm is Q-learning [41], an off-policy method that updates the Q-values using the formula:

$$q^{\pi}(s_t, a_t) \leftarrow q^{\pi}(s_t, a_t) + \alpha(Y - q^{\pi}(s_t, a_t)) \qquad (4)$$

where Y is the temporal difference target and α is the learning rate.

- *Deep Q-Learning (DQN):* Extends Q-learning by using neural networks to approximate Q-values, effective in high-dimensional state spaces [42].
- *Double Q-Learning:* Mitigates overestimation of action values in standard Q-learning [43].
- *Prioritized Experience Replay:* Enhances DQN by prioritizing the replay of critical experiences [44].

In AD, DQN has been applied to highway lane-change decisions where the discrete action space (stay, change left, change right) fits naturally. [45]

*2. Policy-Based Reinforcement Learning (RL):* Policy-based RL directly learns a policy that maps states to actions to maximize cumulative rewards. Unlike value-based RL, it focuses on optimizing a parameterized policy $\pi\theta$ to maximize the expected return:

$$max_{\theta} J(\theta) = max_{\theta} E_{\pi\theta}[G_t \mid S_t = S] \qquad (5)$$

While value-based methods use indirect mechanisms to deduce actions, policy-based RL inherently explores through stochastic policies. These methods tend to be more stable and offer smoother updates.

Challenges in policy-based RL include unstable gradient updates that can lead to poor performance. Methods like:

- Trust Region Policy Optimization (TRPO) [46]: Restrains policy updates using the Kullback-Leibler (KL) divergence, ensuring stability.
- Proximal Policy Optimization (PPO) [47]: Introduces clipping in the surrogate objective function, inspired by gradient clipping, to avoid excessive policy deviations. PPO is widely used in CARLA-based training because its stable updates suit the long episode lengths typical of driving tasks. [48]

*3. Actor-Critic Reinforcement Learning (RL):* Actor-Critic methods combine the strengths of policy- and value-based RL. They include two components:

- Actor: Acquires a behavioral policy establishing a functional mapping from environmental state representations to corresponding action outputs, with the objective of maximizing the expected cumulative reward signal.
- Critic: Approximates the utility of individual states or state-action pairs and furnishes evaluative feedback to progressively refine and optimize the actor's behavioral policy.

The Advantage Function is an essential element in this approach, as it helps to measure how much better or worse an action is compared to the average action in a given state. This helps guide the actor's updates. An example is the Deterministic Policy Gradient (DPG) Algorithm [49], DPG derives deterministic policies, requiring fewer samples in large action spaces, as it integrates only over state spaces.

Actor-Critic methods are preferred for continuous control outputs (steering angle, throttle) in AD because they handle continuous action spaces naturally, unlike value-based methods. [50]

*c) Teacher-Student Paradigm*

Contemporary state-of-the-art methodologies in End-to-End autonomous driving frequently adopt the teacher-student paradigm as a foundational architectural framework. Within this paradigm, a Teacher model, such as CARLA [51], leverages privileged information encompassing ground-truth data pertaining to surrounding agents and cartographic elements to acquire a robust and generalizable driving strategy. The Student model, conversely, is constrained to raw sensory input data, employing behavior cloning as the primary mechanism to approximate and replicate the Teacher's learned behavioral policy. As demonstrated in [52], researchers decomposed the inherently complex tasks of perception and planning into two distinct sequential stages. In the initial stage, a privileged agent was trained utilizing comprehensive environmental data, including the spatial configurations of traffic participants and ground-truth topological layout [52]. This privileged agent circumvents the necessity for visual interpretation by directly accessing the environment's underlying state representation, effectively exploiting information unavailable to standard perception systems [52]. In the subsequent stage, this privileged agent assumes the role of a Teacher, supervising the training of a sensorimotor agent that operates exclusively upon vision-based inputs, devoid of access to any privileged environmental information [52]. This sequential two-stage training procedure ensures that the student agent prioritizes the development of robust sensorimotor learning capabilities, while concurrently benefiting from the supervisory guidance furnished by the Teacher model [52].

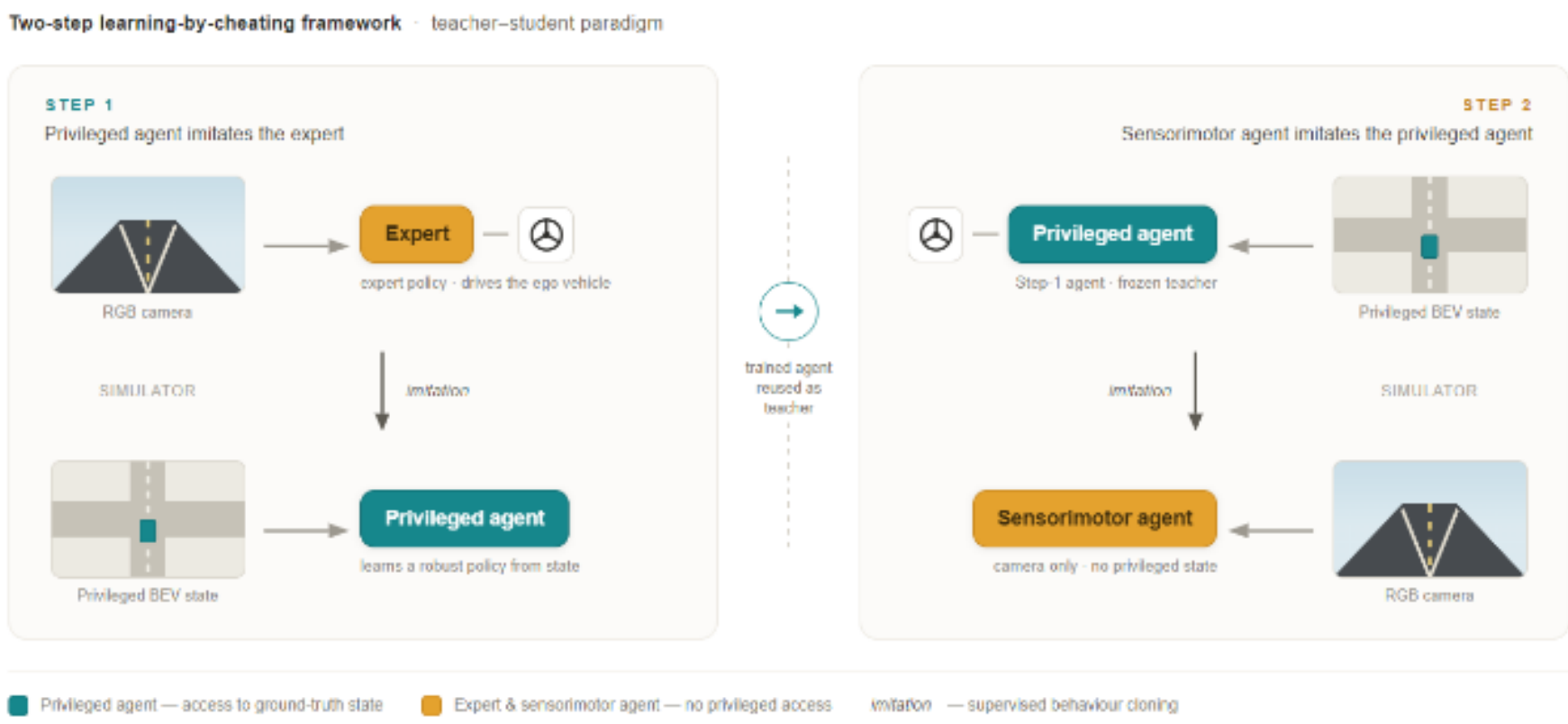

**FIGURE 5 OVERVIEW OF THE TWO-STEP PROCESS IN LEARNING BY CHEATING (TEACHER STUDENT PARADIGM) [52]. (LEFT) A PRIVILEGED AGENT WITH DIRECT ACCESS TO THE ENVIRONMENT STATE LEARNS A ROBUST POLICY BY IMITATING THE EXPERT. (RIGHT) A SENSORIMOTOR AGENT WITHOUT ACCESS TO PRIVILEGED INFORMATION LEARNS THE TASK BY IMITATING THE TRAINED PRIVILEGED AGENT**

However, as highlighted in [53], A prominent limitation inherent to the prevailing Teacher-Student paradigm resides in the requirement for the Student model to autonomously develop an independent planning head a process rendered particularly challenging by the noisy and redundant characteristics of raw sensory input data, compounded by the causal confusion phenomena intrinsically associated with behavior cloning. The proposed approach advocates for the exploitation of the Teacher model's superior planning capabilities, while simultaneously permitting the Student model to concentrate its representational capacity upon perceptual task acquisition.

Despite advancements in designing robust Teacher models and transferring knowledge efficiently, the teacher-student paradigm still faces challenges, particularly in addressing inefficiencies in the knowledge distillation process.

*d) End to end Autonomous driving with the auxiliary task*

The acquisition of control outputs or waypoint predictions for planning of trajectory in autonomous driving is regarded as a highly intricate challenge, attributable to the substantial dimensionality of data parameters inherent in multi-modal sensor inputs. To mitigate this complexity, researchers have devised auxiliary task formulations to augment and support End-to-End learning frameworks. As elaborated in [53], perception and planning modules are trained independently across two sequential stages to reduce overall computational and representational complexity [53]. Uber ATG proposed a more architecturally distinct methodology in their investigation [9], introducing an interpretable End-to-End motion planning framework. Their system employs a backbone model that ingests LiDAR point cloud data and cartographic representations as primary inputs [9]. The resultant outputs encompass bounding box predictions encoding the prospective spatial positions of surrounding traffic participants, constituting the perceptual component, alongside a cost volume parameterized by T filters designated for trajectory planning [9]. A dedicated sampling module generates a diverse set of candidate trajectory proposals, whereupon the associated cost of each proposal is quantified through evaluation against the computed cost volume [9]. The candidate trajectory yielding the minimum aggregated cost is subsequently designated as the definitive planning output (as illustrated in Figure 6).

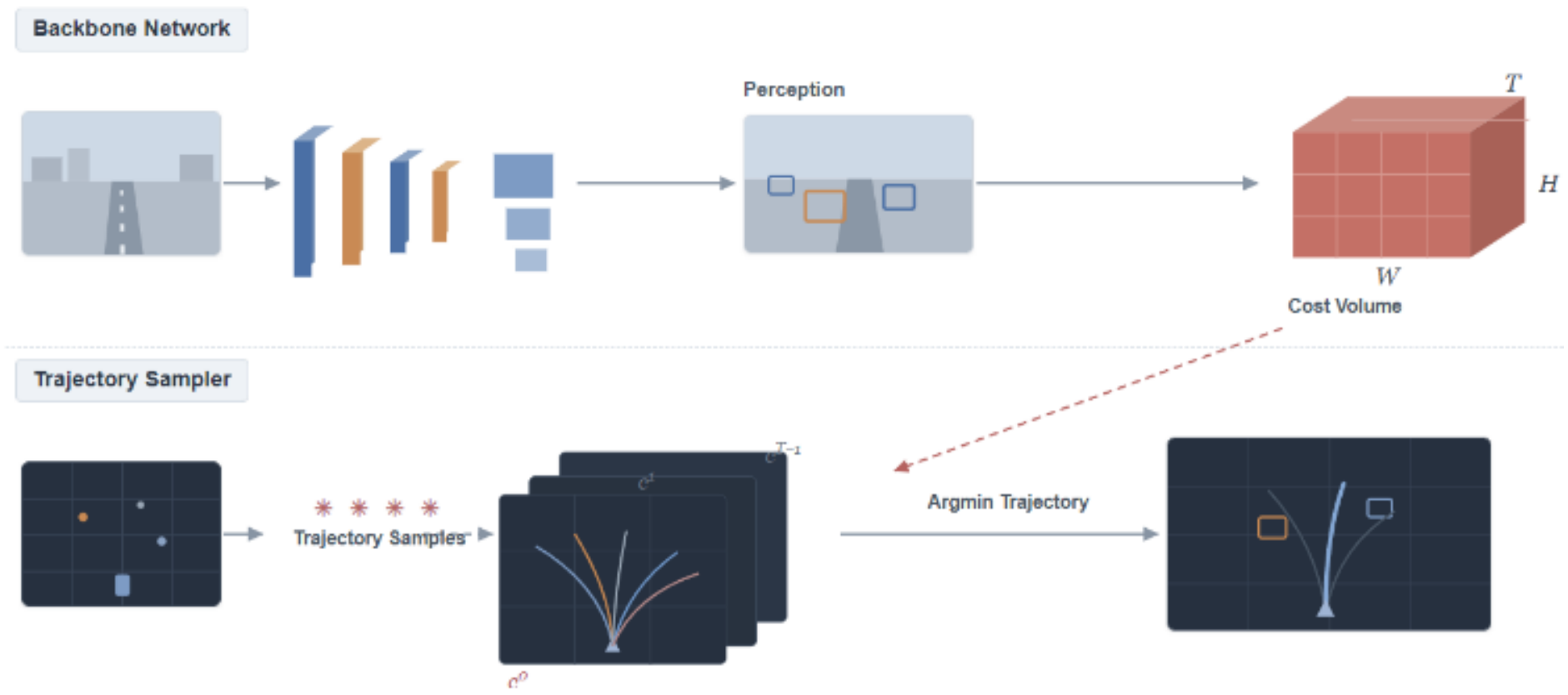

FIGURE 6. END-TO-END AUTONOMOUS DRIVING SYSTEM WITH AUXILIARY TRAINING WITH PERCEPTION LOSSES. [9]

To train this system, a multi-task approach is employed, involving detection, motion forecasting, and supervision from human-driven trajectories. Interestingly, there is no direct supervision for the cost volume. Instead, the network uses a max-margin loss to distinguish good trajectories from bad ones effectively. The overall loss function is:

$$L = L_{perception} + \beta L_{planning} \qquad (6)$$

This multi-task loss formulation enables the model to extract semantically meaningful feature representations, yielding interpretable outputs that are indispensable for diagnosing system failures and enhancing overall operational reliability. The perceptual loss component integrates classification loss, regression loss, and Intersection over Union (IoU) loss, collectively applied to bounding box predictions. The planning loss adopts a max-margin optimization strategy, designating the ground-truth trajectory as a positive exemplar while imposing elevated cost penalties upon randomly sampled negative trajectory candidates. Furthermore, a closely related methodology proposed in [54] incorporates segmentation masks, monocular depth estimation, and optical flow as intermediate representational predictions, serving as auxiliary supervisory signals to facilitate End-to-End control output acquisition via imitation learning [54]. These intermediate auxiliary tasks provide structured inductive guidance to the network, thereby rendering the overarching End-to-End learning process more resilient and amenable to interpretability [54].

*D. Hybrid Architecture*

To overcome the limitations of a standalone modular and end-to-end architecture, one uses a hybrid approach that combines both these approaches. Modular systems work well in well-defined situations where it is important for components to be interpretable. However, they can become too complex and cause error propagation. End-to-end systems, on the other hand, achieve a simpler architecture and exhibit good adaptability to dynamic scenarios, but also lack transparency and require large-scale data [15].

The proposed hybrid architecture merges these two approaches, achieving both interpretability and adaptability. Specifically, the modular components take care of perception and control tasks, while end-to-end methods perform path planning driven by data. Key features include:

- Modular Perception: Uses lidar, stereo camera, and other sensors to accurately detect the surrounding obstacles, traffic lights, and other objects.
- End-To-End Path Planning: Fuses sensor data into (BEVFusion) [15], which is processed by a CNN-based planner to generate dense waypoints.
- Strengths of Hybrid Design [15].
    - Strengthened reliability due to modular handling of sensing and control.

- Used data-driven path planning for mapless environments to enhance flexibility and adaptability.
- It performs balance on map-based and dynamic, unknown scenarios.

The architecture had great success in the 2023 CARLA Autonomous Driving Challenge, performing well in both SENSOR (mapless) and MAP tracks, showing its effectiveness and versatility. The system's modular design allowed debugging per component, while the data-driven components made real-time decisions under dynamic conditions, thus solving autonomous navigation problems robustly [15].

The unique hybrid model is an example of how modular interpretability can be combined with end-to-end adaptivity to advance automated driving systems.

## III. DISCUSSION

### *A. Comparative Overview*

It is important to distinguish two orthogonal classification axes in autonomous driving system design. The first axis concerns environment modeling strategy: model-based approaches (e.g., MPC, A*) rely on explicit mathematical representations of the environment, while model-free approaches (e.g., DRL, imitation learning) learn directly from data. The second axis concerns system architecture: modular systems decompose the driving task into discrete components, while end-to-end systems unify perception and control. These axes interact, a modular system may use model-free methods in its planning component and understanding both is necessary for a complete comparative analysis.

The selection of an appropriate architecture for an autonomous driving system is rarely straightforward. A system designed for highway cruise control operates under fundamentally different constraints than an urban robotaxi, yet both are autonomous driving systems. To address this complexity, this paper introduces a Four-Dimensional Architecture Selection Framework (Figure 7), which provides a structured, repeatable method for selecting between Modular, End-to-End Imitation Learning, Reinforcement Learning, and Hybrid architectures based on the specific requirements of a given application.

The framework organises twelve decision criteria across four dimensions, each targeting a distinct aspect of system requirements. For each criterion, a practitioner answers a binary question yes or no and each answer generates a signal favouring one or more architecture paradigms. The architecture accumulating the highest signal count across all twelve criteria is recommended as the most suitable choice.

**Dimension 1: Safety and Regulatory Compliance** is evaluated first because safety requirements are the hardest constraint in any real-world deployment. It asks three questions: whether full decision traceability is legally required (e.g., under ISO 26262 or UNECE WP.29), whether engineers must be able to isolate a failure to a specific component, and whether formal safety integrity certification such as ASIL-C or ASIL-D is mandated. If any of these apply, the framework signals strongly toward Modular or Hybrid architectures, since end-to-end systems by mapping inputs to outputs through a single opaque neural network cannot natively provide the module-level accountability that certification bodies require.

**Dimension 2: Operating Environment** examines the nature of the deployment scenario. It asks whether the driving domain is highly structured (such as a motorway or a geofenced campus), whether rare or unexpected situations arise frequently, and how dense and unpredictable the surrounding traffic and pedestrian activity is. Structured, predictable environments suit Modular architectures because rule-based planning excels when the world behaves as expected. Unstructured, high- density environments where edge cases are common and agent behavior is erratic favour End-to-End or Hybrid systems, which learn flexible representations from data rather than relying on pre-programmed rules.

**Dimension 3: Data and Computational Resources** addresses the practical constraints on training and running the system. Three factors are assessed: whether a large, labelled training dataset exists for the target domain, whether a high-fidelity driving simulator is available for training, and how much onboard computing power is available at inference time. End-to-end imitation learning demands large-labelled datasets; reinforcement learning requires a capable simulator. Both approaches are computationally expensive at inference. If any of these resources are scarce, the framework signals toward Modular architectures, which can leverage pre-trained components and run efficiently on embedded hardware.

**Dimension 4: Deployment and Operational Context** considers the lifecycle and team context of the project. It evaluates whether the system is a certified production deployment or a research prototype, whether individual components will need to be updated independently over time, and whether the engineering team has cross-functional expertise in both classical control and deep learning. Production systems with strict update processes benefit from modularity, since a single perception model can be swapped without revalidating the full pipeline. Hybrid architectures require expertise in both paradigms and are therefore better suited to teams that span both disciplines.

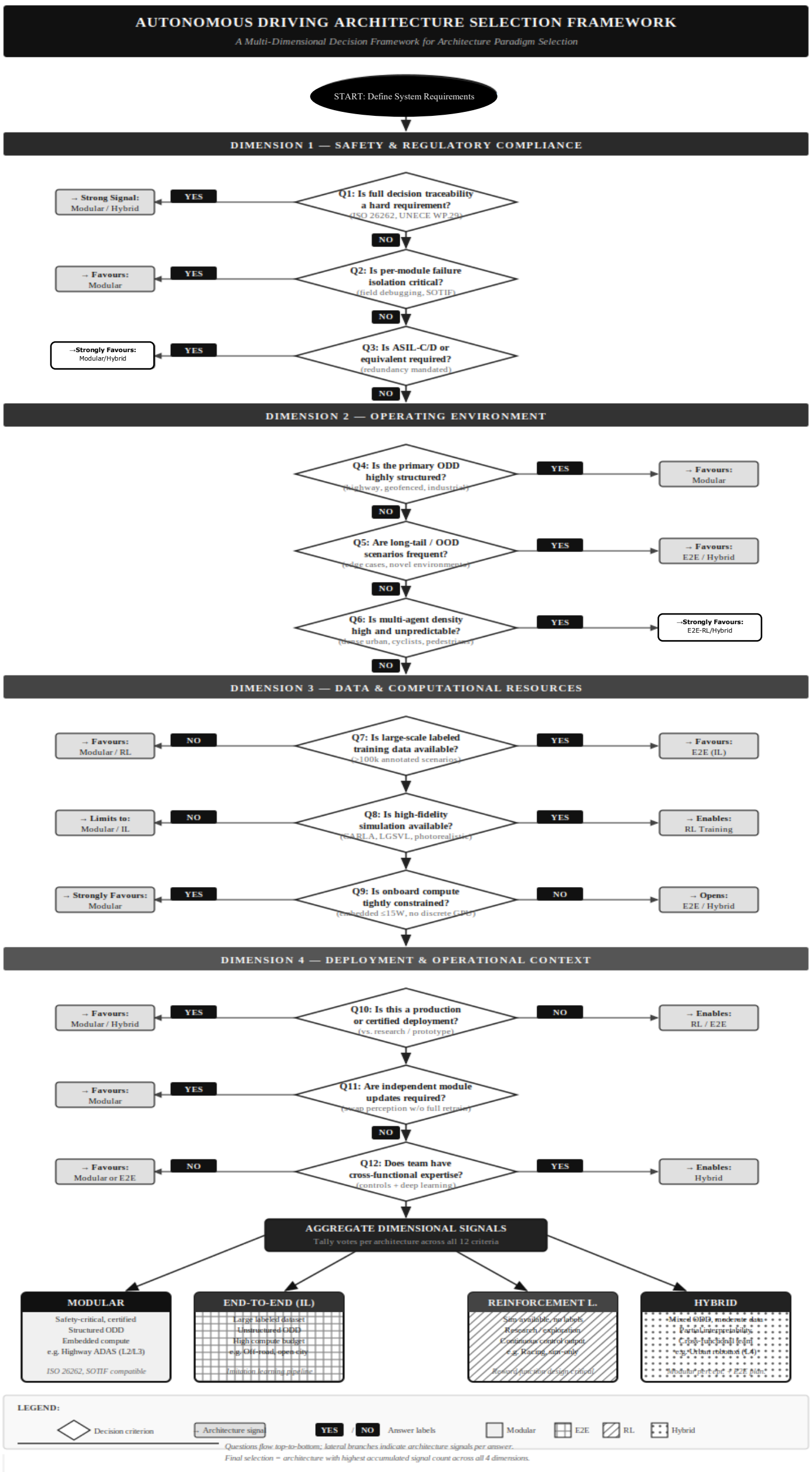

**FIGURE 7. FOUR-DIMENSIONAL ARCHITECTURE SELECTION FRAMEWORK FOR AUTONOMOUS DRIVING SYSTEM DESIGN**

### TABLE 1. COMPARATIVE ANALYSIS OF REPRESENTATIVE AUTONOMOUS DRIVING SYSTEM ARCHITECTURES

**Signal coding - High / Small / None  Medium / Moderate  Low / Large   Row shading - ■ Modular ■ E2E–IL ■ E2E–RL ■ Hybrid**

*Systems are grouped by architecture paradigm (Modular, End-to-End Imitation Learning, End-to-End Reinforcement Learning, Hybrid). Interp. = Interpretability; Data Req. = Training data requirement; Cert. Ready = readiness for safety certification (ISO 26262/SOTIF). DS = Driving Score; RC = Route Completion; LB = CARLA Leaderboard. Note: metrics are not directly comparable across rows due to differing evaluation protocols*

| System / Model | Year | Architecture | Sensors / Input | Benchmark | Key Metric / Result | Interpretability | Data Requirement | Cert. Ready |
|---|---|---|---|---|---|---|---|---|
| **Boss (CMU / DARPA Urban Challenge)** | 2007 | **Modular** | Camera, LiDAR, Radar, GPS/IMU | DARPA Urban Challenge (real-world) | 1st place; completed 60-mile urban course [5] | **High** | **Small** | **High** |
| **Apollo (Baidu)** | 2017 | **Modular** | Camera, LiDAR, Radar, GPS/IMU | Real-world deployment (China) | L4 commercial deployment; >10M km driven | **High** | **Large** | **High** |
| **ALVINN (Pomerleau)** | 1989 - 1992 | **E2E - IL** | Camera (30×32 px), Laser range finder | CMU NAVLAB road-following (real-world) | Successful road-following; up to 70 mph (112 km/h) on highways in 1992 tests | **Low** | **Small** | **Low** |
| **NVIDIA PilotNet** | 2016 | **E2E - IL** | Single front-facing RGB camera | Real-world roads (New Jersey / Nevada) | ~98% autonomy; 0 interventions on 10-mile highway test | **Low** | **Moderate** | **Low** |
| **TransFuser (PAMI version)** | 2022 | **E2E - IL** | Camera (multi-view) + LiDAR BEV fusion | CARLA Leaderboard 1.0; Longest6 | Leaderboard 1.0 DS: 61.18; Longest6 DS: 76.18; −48% collisions vs. geo-fusion baseline | **Low** | **Large** | **Low** |
| **UniAD** | 2023 | **E2E - IL** | 6× Camera (360°) + LiDAR | nuScenes (open-loop planning) | L2 avg: 1.03 m; Collision rate: 0.31%[4] | **Medium** | **Large** | **Low** |
| **LAV (Learning from All Vehicles)** | 2022 | **E2E - IL** | Camera + LiDAR (multi-vehicle data) | CARLA Leaderboard 1.0 | DS: 61.85; RC: 94.46; Won 2021 CARLA Challenge | **Low** | **Large** | **Low** |
| **Roach (RL Coach)** | 2021 | **E2E - RL** | Privileged BEV semantic maps (simulation) | CARLA offline Leaderboard (public routes) | upper-bound on CARLA Leaderboard public routes at publication; widely used as IL teacher - student agents trained under Roach reach 78% success / 88% DS on NoCrash-dense | **Low** | **None(Labelled);High (Simulated Interaction)** | **Low** |
| **Uber ATG NMP (Zeng et al.)** | 2019 | **Hybrid** | LiDAR point cloud + HD Map | Internal Uber ATG real-world dataset (Toronto) | Outperforms separate modules; interpretable cost-volume output | **Medium** | **Large** | **Medium** |
| **CaRINA Agent (Rosero et al.)** | 2023 | **Hybrid** | LiDAR + Stereo Camera + GPS/IMU | CARLA Challenge 2023 (LB1 & LB2) | DS: 41.56 (MAP); RC: 86.03; 1st place SENSORS track LB2 | **Medium** | **Moderate** | **Medium** |

[4] nuScenes open-loop metrics are protocol-dependent (UniAD vs. ST-P3 averaging)

TABLE 2. TWELVE BINARY CRITERIA ORGANIZED ACROSS THE FRAMEWORK'S FOUR DIMENSIONS

| # | Criterion | YES, signals toward |
|---|---|---|
| *C1* | Is full decision traceability legally required (ISO 26262 / UNECE WP.29)? | Modular, Hybrid |
| *C2* | Must engineers be able to isolate a failure to a specific component? | Modular |
| *C3* | Is formal safety certification (e.g., ASIL-C/D) mandated? | Modular, Hybrid |
| *C4* | Is the operating domain highly structured (motorway, geofenced site)? | Modular |
| *C5* | Do rare / long-tail situations arise frequently in this domain? | E2E, Hybrid |
| *C6* | Is surrounding traffic / pedestrian activity dense and unpredictable? | E2E, Hybrid |
| *C7* | Does a large, labelled training dataset exist for this domain? | E2E–IL |
| *C8* | Is a high-fidelity simulator available for training? | E2E–RL |
| *C9* | Is onboard inference compute tightly constrained? | Modular |
| *C10* | Is this a certified production deployment (vs. research prototype)? | Modular, Hybrid |
| *C11* | Will individual components need independent updates over time? | Modular |
| *C12* | Does the team have cross-functional expertise (classical control + DL)? | Hybrid |

**Scoring convention:** a "Yes" response is credited to every architecture named in that criterion's "signals toward" column — so criteria naming two paradigms (e.g., C1, C3, C10: "Modular, Hybrid") contribute to both. Criteria C5 and C6 signal "E2E" at the paradigm-family level rather than a specific training methodology and are therefore credited only to Hybrid in Table 4; the distinction between End-to-End Imitation Learning and End-to-End Reinforcement Learning is instead resolved by the subtype-specific criteria C7 and C8.

By working through these four dimensions sequentially, a practitioner arrives at a recommendation that reflects the real constraints of their application rather than a general preference for one paradigm. Importantly, no single architecture dominates across all criteria: Modular systems excel in safety-critical, resource-constrained deployments; End-to-End systems shine in data-rich, unstructured environments; Reinforcement Learning suits simulation-heavy research settings; and Hybrid systems offer the best balance when moderate interpretability and adaptability are both required. Table 3 and Table 4 reinforces this by showing how ten published systems from ALVINN in 1989 to UniAD in 2023 map onto the framework, confirming that architecture choice has always been driven by application context rather than algorithmic superiority alone

TABLE 3. RESPONSE TO EACH OF THE TWELVE CRITERIA FOR EACH OF THE TEN SYSTEMS IN TABLE 1

| System | C1 | C2 | C3 | C4 | C5 | C6 | C7 | C8 | C9 | C10 | C11 | C12 |
|---|---|---|---|---|---|---|---|---|---|---|---|---|
| *Boss (CMU/DARPA)* | **Y** | **Y** | N | **Y** | N | N | N | N | **Y** | N | **Y** | N |
| *Apollo (Baidu)* | **Y** | **Y** | **Y** | **Y** | N | **Y** | **Y** | **Y** | N | **Y** | **Y** | **Y** |
| *ALVINN* | N | N | N | **Y** | N | N | N | N | **Y** | N | N | N |
| *NVIDIA PilotNet* | N | N | N | **Y** | N | N | N | N | N | N | N | N |
| *TransFuser* | N | N | N | N | **Y** | **Y** | **Y** | **Y** | N | N | N | N |
| *UniAD* | N | N | N | N | **Y** | **Y** | **Y** | N | N | N | N | N |
| *LAV* | N | N | N | N | **Y** | **Y** | **Y** | **Y** | N | N | N | N |
| *Roach* | N | N | N | N | **Y** | **Y** | N | **Y** | N | N | N | N |
| *Uber ATG NMP* | **Y** | **Y** | N | N | **Y** | **Y** | **Y** | N | N | N | **Y** | **Y** |
| *CaRINA Agent* | N | **Y** | N | N | **Y** | **Y** | N | **Y** | N | N | **Y** | **Y** |

TABLE 4. AGGREGATED SIGNAL COUNTS PER PARADIGM, FRAMEWORK PREDICTION, AND DEPLOYED ARCHITECTURE

| System | Mod. | E2E-IL | E2E-RL | Hybrid | Predicted | Actual | Result |
|---|---|---|---|---|---|---|---|
| **Boss** | 5 | 0 | 0 | 1 | Modular | Modular | **Match** |
| **Apollo** | 6 | 1 | 1 | 5 | Modular | Modular | **Match** |
| **ALVINN** | 2 | 0 | 0 | 0 | Modular (weak) | E2E–IL | **Mismatch** |
| **NVIDIA PilotNet** | 1 | 0 | 0 | 0 | Modular (weak) | E2E–IL | **Mismatch** |
| **TransFuser** | 0 | 1 | 1 | 2 | Hybrid | E2E–IL | **Partial** |
| **UniAD** | 0 | 1 | 0 | 2 | Hybrid | E2E–IL | **Partial** |
| **LAV** | 0 | 1 | 1 | 2 | Hybrid | E2E–IL | **Partial** |
| **Roach** | 0 | 0 | 1 | 2 | Hybrid | E2E–RL | **Mismatch** |
| **Uber ATG NMP** | 3 | 1 | 0 | 4 | Hybrid | Hybrid | **Match** |
| **CaRINA Agent** | 2 | 0 | 1 | 3 | Hybrid | Hybrid | **Match** |

The framework correctly recommends the actual deployed architecture either as an exact match or as a partial match for seven of the ten published systems. Four systems show an exact match between predicted and actual paradigm: Boss, Apollo, Uber ATG NMP, and CaRINA Agent. In each of these cases the leading signal count under Table 2's criteria coincides with the paradigm the system's developers actually deployed.

Three systems (TransFuser, UniAD, and LAV) show partial matches: the framework signals Hybrid as the strongest single paradigm, while all three systems are implemented as End-to-End Imitation Learning models that incorporate auxiliary modular-style supervision (e.g., BEV fusion, intermediate perception losses), placing them architecturally close to the Hybrid category as defined in this paper even though they are conventionally labelled End-to-End in the literature.

Three systems are mismatches, of two distinct kinds. ALVINN (1989) and NVIDIA PilotNet (2016) are early-generation research demonstrations rather than deployment-oriented systems, and none of the twelve criteria are strongly triggered for narrow-scope demonstration platforms: there is no certification pressure, no production deployment requirement, and, in ALVINN's case, no large training corpus by contemporary standards. Consequently, the framework returns only a weak Modular signal by default (through low data availability and unconstrained compute criteria) rather than a confident recommendation in either direction. This indicates that the framework is best suited to deployment-stage architecture decisions and is not intended to retrospectively classify early research prototypes that predate the deployment and certification considerations the framework's criteria are built around. We treat this as an explicit scope boundary of the framework rather than a failure of its underlying logic.

Roach is a mismatch of a different kind. The framework signals Hybrid as the strongest paradigm (2 Hybrid-aligned criteria against 1 E2E–RL-aligned criterion, driven by C5 and C6's dense-traffic / long-tail signal), while Roach was actually deployed as a pure End-to-End Reinforcement Learning system trained against privileged BEV state in simulation, with no modular or hybrid components in its final architecture. Unlike TransFuser, UniAD, and LAV, Roach's design does not incorporate auxiliary modular-style supervision, so the "architecturally close to Hybrid" partial-match reasoning does not extend to it; it is recorded here as a straightforward mismatch rather than a partial match. Note that Roach's 2-to-1 signal split is a clear plurality for Hybrid, not a tie, so no tie-break convention is invoked for this system.

Ties are broken according to the following convention: when two or more paradigms receive an equal signal count, precedence is given first to any paradigm signalled by Dimension 1 (Safety and Regulatory Compliance) criteria, and second to the paradigm matching the system's evaluation context (e.g., simulation-only vs. real-world deployment) as reported in Table 1. No system in the current ten-system validation set produces a tie under the tallies in Table 4.

*a) Environment Modeling Paradigms: Model-Based vs. Model-Free*

*1.* Model-Based Approaches

i. *Autonomous Model Predictive Control (MPC):* Model Predictive Control is a well-known model-based approach used in autonomous driving. It predicts future vehicle states based on a dynamic vehicle model and its environment. By optimizing a trajectory over a finite horizon, MPC ensures safe and efficient path planning. Example: Google's Self-Driving Car project utilized an MPC framework to handle complex vehicle dynamics and optimize control signals for smooth driving [55].

ii. *Trajectory Optimization Systems:* Techniques like D* and A* algorithms are employed to compute optimal paths based on graph representations of the road network. These methods rely on predefined models of the environment and vehicle motion [56]. Example: Tesla's Autopilot uses model-based planning to anticipate nearby vehicle movements and plan optimal trajectories.

iii. *Physics-Based Vehicle Dynamics Models:* These approaches use kinematic and dynamic models of vehicles to predict motion and ensure stability during maneuvers. For instance, methods that model tire-road interactions or lateral and longitudinal dynamics [57]. Example: Autonomous race cars in the Roborace series implement such physics-based models for high-speed control.

2. Model-Free Approaches

i. *Deep Q-Learning for Path Planning*: This reinforcement learning algorithm trains an agent to select optimal actions without a predefined model. By interacting with the environment and optimizing rewards, it learns an effective driving policy. Example: Researchers have developed Deep Q-Network (DQN) models for tasks like highway lane changes and obstacle avoidance. [41] [42]

ii. *Proximal Policy Optimization (PPO):* PPO is a policy gradient-based reinforcement learning algorithm often used in model-free settings to teach autonomous vehicles driving behaviors directly from simulated environments. Example: OpenAI's CARLA-based simulation experiments demonstrate the effectiveness of PPO in handling intersections and pedestrian crossings. [47]

iii. *Imitation Learning:* Using data-driven models, such as behavior cloning, these approaches learn to mimic expert driving demonstrations without relying on explicit environment models. Example: Nvidia's PilotNet, which maps camera input to steering commands, is a classic example of model-free imitation learning. [11] [12]

*b) System Design Paradigms: Modular vs. End-to-End Learning*

*1.* Modular Architecture

One of the primary benefits of modular systems is their interpretability. By dividing tasks into distinct components, such as perception, prediction, planning, and control, it becomes easier to trace errors and debug specific modules. This clarity is especially important when diagnosing issues during system development and testing [58] [16].

Another significant advantage is their adaptability. Modular systems allow individual components to be updated or replaced independently, enabling the integration of new technologies or the optimization of specific functions. This modularity provides flexibility, making it possible to enhance parts of the system without requiring a complete overhaul [59] [50].

Reliability is also a key strength of modular architectures. These systems often employ well-established techniques, such as Model Predictive Control (MPC), which are known for their robust performance in structured and predictable driving environments. This ensures that the system can handle well-defined scenarios effectively [59].

However, modular systems face challenges, particularly with integration complexity. Coordinating smooth transitions and communication between modules can be difficult, especially when assumptions made by different components are not aligned. This misalignment can lead to inefficiencies or errors in the overall system [50]. Another drawback is error propagation. A mistake in one module, such as incorrect object detection in the perception component, can negatively affect downstream processes, including planning and control. This cascading effect reduces the overall effectiveness and reliability of the system [50] [60].

Finally, modular systems often struggle with limited generalization. These systems frequently rely on handcrafted features, which can make them less effective in dynamic and unstructured environments. Their ability to adapt to diverse scenarios is constrained, as they are often tailored for specific, predictable situations [60].

*2.* End-to-End Learning

One of the key strengths of end-to-end systems is their simplified pipeline. By directly mapping raw sensor inputs to control outputs, these systems eliminate the need for intermediate steps, complex module interactions, and handcrafted rules. This streamlined design reduces architectural complexity and minimizes dependencies between components, making them easier to implement in some scenarios [60].

End-to-end systems also offer significant generalization potential. With access to diverse and representative training data, these systems can adapt to a wide range of environments and tasks, including those not explicitly pre-programmed. This flexibility makes them suitable for dynamic and unpredictable real-world conditions [59] [9]. Another major advantage is their data-driven optimization. End-to-end models improve their performance as the size and diversity of training datasets increase. They can learn from vast amounts of simulated driving data or human demonstrations, enabling continuous refinement of their decision-making capabilities [50] [60] [9].

Despite their advantages, end-to-end systems face challenges, primarily due to their low interpretability. The black-box nature of deep learning models makes it difficult to understand the rationale behind their decisions. This lack of transparency complicates debugging, troubleshooting, and ensuring compliance with regulatory standards [9] [58].

Another drawback is their high data demands. Training these systems requires extensive labeled datasets that capture diverse scenarios, including rare edge cases like accidents or adverse weather conditions. Such data can be difficult and expensive to collect [50]. End-to-end systems are also prone to overfitting risks. If the training data is not diverse enough, the model may become too specialized and perform poorly in unfamiliar or novel situations. This limits the system's reliability across different environments [60] [9].

Finally, there are safety concerns associated with end-to-end architectures. Without explicit intermediate checks or redundancy, ensuring robustness and reliability in critical situations is a significant challenge. This lack of built-in safeguards can lead to failures in high-stakes scenarios [50] [60].

### *B. Trade-offs:*

#### a) Transparency vs. Flexibility

The components of modular architectures are interpretable, and therefore they are transparent systems. It is important for safety-critical applications like autonomous driving that ensure accountability and aid in failure analysis. On the other hand, end-to-end systems are more flexible and can better adapt to complex, unstructured situations through direct learning from data. However, it is hard to understand why and how constantly high-performance CNNs fail. Thus, there is a significant trade-off between performance and durability. [13] [10] [15].

#### b) Scalability

End-to-end models have demonstrated scalability in simulation-based settings, where increasing dataset size consistently improves performance [61], though real-world scalability remains constrained by data collection costs and domain gap challenges, and to adapt to different driving scenarios [14]. The ability to adapt behavioral policies to new driving contexts improves task performance. Since the environment is predictable and structured, modular systems' pre-defined components of the architecture can perform well without going through a laborious retraining process to new data. [13] [16] [8].

#### c) Error Diagnosis

Due to the independence of components in modular systems, developers can identify errors in a particular module like perception or planning and subsequently fix it. This makes fixing bugs easier and quicker. On the other hand, the end-to-end systems are interconnected so debugging most likely has to be done in its entirety. The fault may not lie in the final decision but instead indicates there was an error in the input or any of the intermediate representations. Fixing solutions in end-to-end systems becomes complex because of this. [8] [13].

### *C. Future Trends:*

In autonomous driving, hybrid architectures combine modular and end-to-end learning approaches, leveraging the strengths of both to achieve superior results. Most of the time, architectures make use of modules for tasks that necessitate interpretability and precision, e.g., perception and localization, while making use of end-to-end learning for those modules which benefit from data, e.g., motion planning and control. A mix and hybrid model can use a modular perception pipeline and end-to-end planner. The steering is then learned from pre-processed sensor data. This combination helps tackle the challenges of interpreting end-to-end learning and improving scalability in modular systems while striking a balance between accuracy, adaptability, and computational efficiency. For example, Uber ATG's interpretable motion planner combines the use of end-to-end learning for perception as well as motion forecasting with modular trajectory optimization algorithms. This combination leads to a balance between data-driven learning and interpretable decision-making [58].

These systems enhance generalization while maintaining interpretability, enabling safer and more robust autonomous driving solutions.

## IV. CONCLUSION:

The continued deployment of autonomous vehicles introduces several open research challenges. One of these challenges is the selection of an appropriate system architecture. Modular system design offers transparency, adaptability, and reliability. by decomposing complex tasks into well-defined sub-problems, they can be debugged easier, adapted to newer technology faster, and most importantly, they perform more robustly in a structured environment. Though modular systems are now more commonly employed, they face complex issues involving integration problems, error propagation, limited generalization capabilities, and so on. These issues are evident mainly in dynamic, unstructured environments. On the other hand, end-to-end learning systems offer scalability, generalization, and data-driven optimization benefits. These systems are very good at handling different types of data for different types of driving conditions. Still, they are hard to interpret and require large amounts of data with their black-box nature and lack of intermediate checks making them overfit and unsafe. Hybrid architectures in autonomous driving offer one of the most promising research direction by integrating the strengths of both modular and end-to-end approaches. This synergy enables systems to benefit from the interpretability of modular components while leveraging the adaptability of end-to-end learning, leading to enhanced generalization, accuracy, and computational efficiency. To support practitioners in navigating this design space, this paper proposed a Four-Dimensional Architecture Selection Framework built around twelve binary criteria spanning safety and regulatory compliance, operating environment, data and computational resources, and deployment context. Validated against ten published systems deployed between 1989 and 2023, the framework correctly identified the deployed architecture for seven cases four as an exact match and three as a partial match while also clarifying its scope boundary: it is best suited to deployment-stage decisions rather than the retrospective classification of early research prototypes. For example, Uber ATG's interpretable motion planner combines data-driven learning for perception and motion forecasting with modular trajectory optimization algorithms, achieving a balance between flexibility and transparency. Future work should focus on three open problems: establishing standardized cross-benchmark evaluation protocols that fairly compare modular and end-to-end systems; developing hybrid architectures with formally verifiable safety properties; and reducing the sim-to-real gap for end-to-end learned policies through domain randomization and adaptive transfer techniques.

## V. ACKNOWLEDGEMENT:

AI-assisted tools were used during manuscript preparation for language editing and drafting support, under the author's supervision and review; all technical content, analysis, and conclusions are the author's own.